\documentclass{article}\usepackage{RR}

\usepackage{multicol}
\usepackage{ifthen}
\usepackage{amssymb} 
\usepackage{ifthen}
\usepackage{amsmath}
\usepackage{amsfonts}

\RRdate{May 2008}
\RRNo{6527}

\RRauthor{Nikolaus Hansen\thanks[snf]{Adaptive Combinatorial Search Team, 
           Microsoft Research--INRIA Joint Centre.
           28, rue Jean Rostand, 91893 Orsay Cedex, France. 
          email:\texttt{forename.name@inria.fr}. }
}

\authorhead{Nikolaus Hansen}
\RRtitle{Adaptation du Pas Deux-Point dans CMA-ES }
\RRetitle{CMA-ES with Two-Point Step-Size Adaptation}
\titlehead{}
\newcommand{\kommentar}[1]{}
\newcommand{\kom}[1]{}
\newcommand{\com}[1]{}
 \RRabstract{We combine a refined\kom{polish, enhanced, promoted,
 amended, revised, enriched, meliorated} version of \emph{two-point
 step-size adaptation} with the \emph{covariance matrix adaptation
 evolution strategy} (CMA-ES). Additionally, we suggest polished
 formulae for the learning rate of the covariance matrix and the
 recombination weights. In contrast to \emph{cumulative step-size
 adaptation} or to the \emph{1/5-th success rule}, the refined
 two-point adaptation (TPA) does not rely on any internal model of
 optimality. In contrast to conventional \emph{self-adaptation}, the TPA will
 achieve a better target step-size in particular with large
 populations. The disadvantage of TPA is that it relies on two
 additional objective function evaluations. }

\RRkeyword{
           optimization, 
           evolutionary algorithms, 
           covariance matrix adaptation,
           step-size control,
           self-adaptation,
           two-point adaptation
           } 
\RRprojets{Adaptive Combinatorial Search et TAO}
\RRtheme{\THCog} 
\RCSaclay 

\newcommand{\nref}[1]{\nolinebreak[4]\hspace{0.2845em plus0.05em minus0.17em}\nolinebreak[4]\ref{#1}}

\begin{document}

\makeRR   

\newcommand{\klamstri}[1]{(#1\mbox{-)}\linebreak[0]}
\newcommand{\eye}[1]{\textbf{#1}}
\newcommand{\rklam}[1]{\left({#1}\right)}
\newcommand{\N}{\ensuremath{n}}
\newcommand{\Nat}{\mathbb{N}}
\newcommand{\R}{\mathbb{R}}
\newcommand{\Z}{\mathbb{Z}}
\newcommand{\Rn}{\R^{\N}}
\newcommand{\findicator}[1]{1\hspace{-0.35em}1\hspace{-0.1em}_{#1}}
\newcommand{\x}{\ensuremath{\vec{x}}}
\newcommand{\eg}{{\it e.g.}}
\newcommand{\ie}{{\it i.e.}}
\newcommand{\noise}{\ensuremath{N_f}}
\newcommand{\Extext}[1]{{\mathrm{E}}\hspace{-0.0em}\big[#1\big]}
\newcommand{\ccov}{c_{\mathrm{cov}}}
\newcommand{\mucov}{\ensuremath{\mu_{\mathrm{cov}}}}
\newcommand{\F}{\ensuremath{L}}
\newcommand{\Fmean}{\ensuremath{\widehat{L}}}
\newcommand{\SearchSpace}{{\cal S}}
\newcommand{\Rinfty}{\ensuremath{R^\infty}}
\newcommand{\ma}[1]{\mathchoice{\mbox{\boldmath$\displaystyle#1$}}
  {\mbox{\boldmath$\textstyle#1$}} {\mbox{\boldmath$\scriptstyle#1$}}
  {\mbox{\boldmath$\scriptscriptstyle#1$}}}
\renewcommand{\vec}[1]{\mathchoice{\mbox{\boldmath$\displaystyle#1$}}
  {\mbox{\boldmath$\textstyle#1$}} {\mbox{\boldmath$\scriptstyle#1$}}
  {\mbox{\boldmath$\scriptscriptstyle#1$}}}
\newcommand{\SNR}{SNR}
\newcommand{\sigerr}{\sigma_\varepsilon}
\newcommand{\mwl}{\ensuremath{\mu/\mu_\mathrm{\textsc{w}},\lambda}}

\section{Introduction}
 In the Covariance Matrix Evolution Strategy (CMA-ES) \cite{hansen:01}
 two separate adaptation mechanism are performed to determine
 variances and covariances of the search distribution. One for
 (overall) step-size control, a second for adaptation of a covariance
 matrix. The mechanisms are mainly independent and can therefore, in
 principle, be replaced separately. While the standard step-size
 control is \emph{cumulative step-size adaptation} (CSA), also a
 success-based control was successfully introduced for the
 (1+$\lambda$)-CMA-ES in \cite{igel:06b}.

 The CSA has a few drawbacks. 

\begin{itemize}

\item For very large noise levels the
 target step-size becomes zero, while the optimal step-size is still
 positive \cite{Beyer:2003}. 

 \item For large population sizes ($\lambda>10\,n$) the original
 parameter setting seemed not to work properly \cite{hansen:03}---the
 notion of tracking a (long) path history seems not to perfectly mate
 with a large population size (large compared to the search space
 dimension). An improved parameter setting introduced in
 \cite{hansen2004ppsn} shortens the backward time horizon for the
 cumulation and performs well also with large population sizes
 \cite{hansen2004ppsn, auger:2005}.

 \item The expected size for the displacement of the
 population mean under random selection is required. To compute a
 useful measurement independent of the coordinate system, the
 principle axes of the search distribution are needed.  They are more
 expensive to acquire (at least by a constant factor) than a simple
 matrix decomposition that is in any case necessary to sample a
 multivariate normal distribution with given covariance matrix.

 \item Because the length of an evolution path is compared to its
 expected length, the measurement is sensitive to the specific sample
 procedure of new candidate solutions and also, for example, to repair
 mechanisms for solutions. 

\end{itemize} 

 Despite these disadvantages, CSA is regarded as first choice for
 step-size control in the (\mwl)-ES, due to its advantages.
 Nonetheless, the disadvantages rise motivation to search for
 alternatives. Here, we suggest \emph{two-point step-size adaptation}
 (TSA) as one such alternative. 

 Two-point self-adaptation was introduced for backpropagation in
 \cite{salomon1996abt} and later applied in Evolutionary Gradient
 Search \cite{salomon1998eaa}. In evolutionary search, two-point
 adaptation resembles self-adaptation on the population level. The
 principle is utmost simple: two different step lengths are tested for
 the mean displacement and the better one is chosen. In the next
 section, we integrate a slightly refined\kom{polished,enhanced} TSA
 in the CMA-ES and additionally introduce polished formulae for the
 recombination weights and the learning rates of the covariance
 matrix.


\section{The Algorithm: CMA-ES with TPA}
\newcommand{\m}{\ensuremath{\vec{m}}}
\newcommand{\mold}{\m_\text{old}}
\newcommand{\y}{\vec{y}}
\newcommand{\ymean}{{\langle\vec{y}\rangle}}
\newcommand{\il}{{i:\lambda}}
\newcommand{\C}{\ensuremath{\ma{C}}}
\newcommand{\B}{\ma{B}}
\newcommand{\D}{\ma{D}}
\newcommand{\T}{\mathrm{T}}
\newcommand{\Cinv}{\C^{-\frac{1}{2}}}
\newcommand{\Id}{\mathbf{I}}
\newcommand{\mueff}{\ensuremath{\mu_\textsc{w}}}
\newcommand{\cs}{c_\sigma}
\newcommand{\ds}{d_\sigma}
\newcommand{\ps}{\vec{p}_\sigma}
\newcommand{\cc}{c_{\ma{c}}}
\newcommand{\pc}{\vec{p}_{\ma{c}}}
\newcommand{\hsig}{h_\sigma}
\newcommand{\Normal}[1]{{\mathcal N}\hspace{-0.13em}\left(#1\right)}
\newcommand{\NormalNullI}{{\mathcal N}
        \hspace{-0.13em}\left({\vec{0},\Id}\right)}
\newcommand{\chiN}{\ensuremath{\widehat{\chi}_\N}}
\newcommand{\bigO}{{\cal O}}
\newcommand{\defgleich}{\stackrel{\mathrm{def}}{=}}

\newcommand{\TPA}{\ensuremath{\text{TPA}}}
\newcommand{\TPAalphasel}{\ensuremath{\alpha_\text{act}}}
\newcommand{\TPAalphamean}{\ensuremath{\alpha_\mathrm{s}}}
\newcommand{\TPAalpha}{\ensuremath{{\alpha}}}
\newcommand{\alphap}{\ensuremath{{\alpha'}}}
\newcommand{\TPAbeta}{\ensuremath{{\beta}}}
\newcommand{\calpha}{\ensuremath{c_\alpha}}
\newcommand{\old}{\text{old}}

 Our description of the CMA-ES closely follows
 \cite{hansen2006eda,hansen2004ppsn,hansen:08} and replaces CSA with
 TSA. Given an initial mean value $\m\in\Rn$, the initial covariance
 matrix $\C=\Id$ and the initial step-size $\sigma\in\R_+$, the new
 candidate solutions $\x_k$ obey
\begin{equation}\label{eq:basic}
  \x_k = \m + \sigma\, \y_k,\quad\text{for~~}  k=1,\ldots,\lambda
  \enspace,
\end{equation}
where $\y_k\sim\Normal{\vec{0},\C}$ denotes the realization of a
normally distributed random vector with zero mean and covariance
matrix $\C$. The solutions $\x_k$ are evaluated 
and ranked such that $\x_\il$ becomes the $i$-th best solution vector
and $\y_\il$ the corresponding random vector realization.

For $\mu<\lambda$ let
\begin{equation}\label{eq:zmean}
 \ymean = \sum_{i=1}^{\mu}w_i \y_\il,
    \quad w_1\ge\dots\ge w_\mu>0,
    \quad \sum_{i=1}^{\mu}w_i = 1
\end{equation}
be the weighted mean of the $\mu$ best ranked $\y_k$ vectors. The
recombination weights sum to one. The \emph{variance
effective selection mass} is defined as 
\begin{equation}
\mueff = \frac{\left(\sum_{i=1}^{\mu}w_i\right)^2}{\sum_{i=1}^{\mu}w_i^2} 
       = \frac{1}{\sum_{i=1}^{\mu}w_i^2} 
       \ge 1
\enspace.
\end{equation}
 From the definition follows that $1\le\mueff\le\mu$ and $\mueff=\mu$ for
 equal recombination weights. The role of $\mueff$ is analogous to the
 role of the parent number $\mu$ when the recombination weights are
 all equal.  Usually $\mueff\approx\lambda/4$ is
 appropriate. Weighted recombination is discussed in more detail in
 \cite{arnold2006wme}.

The default parameter values are
\begin{equation}\label{eq-def-lam}
\lambda=4+\lfloor3\,\ln\N\rfloor, \quad
\mu'=\frac{\lambda}{2}, \quad
\mu=[\mu'] \quad\text{and}
\end{equation}
\begin{equation}\label{eq-def-w}
w_{i} = \frac{\ln(\mu'+0.5)-\ln i}{\sum_{j=1}^{\mu}(\ln(\mu'+0.5)-\ln j)}
\quad\mbox{for~~}i=1,\dots,\mu
\enspace,
\end{equation}
 where $[\mu']$ denotes the integer value closest to $\mu'$,
 preferably chosing the smaller integer value in case, such that
 $w_{[\mu']}>0$. The first $[0.2\mu']$ weights sum to about
 $0.5$. Conducting restarts with increasing value of $\lambda$
 is a valuable option \cite{auger:2005}.

 In the remainder, the generation step is completed with the updates
 of $\m$, $\sigma$, and $\C$, where two additional state variables,
 $\TPAalphamean\in\R$ and $\pc\in\Rn$, will be introduced and the
 method parameters are discussed in Section\nref{sec:parameters}.

\subsection{The Mean} The distribution mean is updated according to 
\begin{equation}\label{eq:mean}
  \m\gets \m + \sigma \, \ymean \com{=\sum_{i=1}^{\mu}w_i \x_\il}
\enspace.
\end{equation}
%
Given $\sigma$ from Equation \eqref{eq:basic}, Equation
\eqref{eq:mean} can also be written as 
\begin{equation}\label{eq:meanfromx}
\m\gets\sum_{i=1}^{\mu}w_i \x_\il
\enspace. 
\end{equation}

 \subsection{Step-Size Control: Two-Point Adaptation (TPA)} A two-point
 self-adap\-tive scheme is implemented based on
 \cite{salomon1998eaa}. We compute two additional function evaluations
\begin{eqnarray}
   f_{+} &=& f(\m + \alphap\,\sigma\,\ymean) \label{eq:TPAfevalplus}\\
   f_{-} &=& 
              f(\m - \alphap\,\sigma\,\ymean) \label{eq:TPAfevalminus}
\enspace,
\end{eqnarray}
 where $f$ is the objective function to be minimized, \m\ is the new
 (updated) mean value, and $\alphap\approx0.5$ is the
 test width parameter. The factor $\pm\alphap$ in
 the equations is chosen symmetrical about the new mean \m.

 The step-size should increase if $f_+$ is better than $f_-$, and
 decrease otherwise. Using the values $f_+$ and $f_-$ we set
\begin{equation}
   \TPAalphasel = \left\{
      \begin{array}{lll}
      -\TPAalpha + \TPAbeta & < 0, \text{~if $f_-$ is better (smaller) than $f_+$} \\
	\TPAalpha &> 0, \text{~otherwise}
      \end{array}
   \right.
\end{equation}
Initializing $\TPAalphamean=0$, 
the new step-size is calculated according to 
\begin{eqnarray}
\TPAalphamean &\gets& \TPAalphamean + \calpha(\TPAalphasel-\TPAalphamean) 
   \,=\, (1-\calpha)\,\TPAalphamean + \calpha\TPAalphasel 
  \label{eq:ps}\\
\sigma &\gets& \sigma\times\,\exp\rklam{{\TPAalphamean}}
\label{eq:sigma}\end{eqnarray}
 where $1/\calpha\ge1$ determines the backward time horizon for
 smoothing the step-size changes in the
 generation sequence. The default parameter settings are
\begin{equation}\label{eq-def-sig}
\alphap=0.5,\quad \TPAalpha= 0.5, \quad\TPAbeta = 0, \quad
\calpha= 0.3
\enspace.
\end{equation}

\paragraph{Comparison to the previous formulation} 
 The two-point step-size adaptation described here differs from
 \cite{salomon1998eaa} in that the test steps are distinguished from
 the step-size changes by using (i) a \emph{symmetrical} test
 step about the new \m, (ii) different test width and change parameters
 and (iii) a smoothing for the step-size change. Furthermore, the
 \emph{original} step-size is used for updating \m. Setting
 $\alphap=0.8$, $\TPAalpha=\ln(1.8)\approx0.588$, $\TPAbeta=0$,
 $\calpha=1$, replacing $-\alphap$ with
 $-\alphap/(1+\alphap)$ in Equation \eqref{eq:TPAfevalminus} and using
 the new step-size for finally updating the mean \m\, recovers the
 step-size adaptation from \cite{salomon1998eaa}. We do not expect an
 essentially different behavior due to our refinements in most cases.

 Step-size changes are essentially multiplicative.  A factor
 $\exp(\pm\TPAalpha)$ can be used to realize changes of $\sigma$,
 which is symmetrical about $1$ in the log scale.  On the other hand,
 using such factors for generating \emph{test} steps extends the step
 further by $\exp(+\TPAalpha)>1$ than reducing it by
 $\exp(-\TPAalpha)<1$.  Assuming the most simple spherical objective
 function model and \emph{optimal step-size}, where $f(\m + a\ymean)$
 about the new mean \m\ is minimal for $a=0$ and 
$$
  f(\m + a\ymean) = f(\m - a\ymean)
\enspace,
$$ 
 a larger test step 
$$ 
  f(\m_\old + \exp(\alpha'')\sigma\ymean)  
  = f(\m + \alphap\sigma\ymean)
\enspace,
$$
given $\alpha''=\ln(1+\alphap)$, is disfavored compared to 
$$ 
 f(\m_\old +
 \exp(-\alpha'')\sigma\ymean) 
 = 
 f\Big(\m -
 \frac{\alphap}{1+\alphap}\sigma\ymean\Big) 
\enspace.
$$
 The step-size will systematically decrease, the target step-size is
 smaller than the optimal step-size. On simple functions, like the
 sphere model, this effect might well lead to a performance
 improvement, because the optimum can be approached quickly and
 therefore the optimal step-size decreases fast. The sub-optimal
 target step-size ``anticipates'' this change. Nevertheless, in
 general, we tend to favor an agreement of target and optimal
 step-size and therefore we are in favor of symmetrical test
 steps.\footnote{Good algorithm design must at times prefer the
 reasonable to the optimal performance in order to avoid overfitting to
 specific test scenarios. }

 Following \cite{salomon1998eaa}, the update of \m\ in Equation
 \eqref{eq:mean} could be postponed until after the step-size is
 updated in Equation \eqref{eq:sigma} (Equations
 \eqref{eq:TPAfevalplus} and \eqref{eq:TPAfevalminus} must be revised
 accordingly using the old mean value). Whether or not this results in
 a better \m\ cannot be decided without additional costs, because
 neither the original step-size nor the updated step-size are usually
 tested. Furthermore, Equation \eqref{eq:meanfromx} would not hold
 anymore. Empirically, using the new step-size leads to slightly
 higher convergence rates in norm optimization (sphere function) in
 small dimensions.

\newcommand{\cone}{\ensuremath{c_1}}
\newcommand{\cmu}{\ensuremath{c_\mu}}
 \subsection{Covariance Matrix Adaptation (CMA)} The covariance matrix
 admits a rank-one and a rank-$\mu$ update. For the rank-one update an
 evolution path $\pc$ is constructed.
 \begin{eqnarray}
 \pc &\gets& (1-\cc)\,\pc + \hsig\sqrt{\cc(2-\cc)\mueff}\,\ymean
    \label{eq:pc}\\
 \C &\gets& (1-\cone-\cmu)\,\C + \cone
    \underbrace{\pc\pc^\T
           }_{\hspace{-2.50em}\mbox{rank-one update}\hspace{-1.50em}} 
    +\; \cmu 
    \underbrace{\sum_{i=1}^{\mu}w_i\y_\il\y_\il^\T
    \hspace{-0ex}}_{\mbox{rank-$\mu$ update}}
  \label{eq:C}
  \enspace,
 \end{eqnarray}
 where $\hsig= 0$ if $\TPAalphamean > (1-(1-\calpha)^9)
 (1-(1-\calpha)^g)\,\TPAalpha$, and $1$ otherwise, where $g$ is the
 generation counter.  The update of $\pc$ is stalled when
 \TPAalphamean\ is large. The stall is decisive after a change in the
 environment which demands a significant increase of the step-size.
 Fast changes of the distribution shape are postponed until
 after the step-size has increased to a reasonable value.

 For the covariance matrix update, the cumulation in
 \eqref{eq:pc} serves to capture dependencies between consecutive
 steps.  Dependency information would be lost for $\cc=1$, because a
 change in sign of $\pc$ or $\y_\il$ does not matter in \eqref{eq:C}.

The default parameter settings are 
\begin{equation}
\cc=\frac{4}{n+4}, \quad\mucov=\mueff,
\label{eq-def-b}
\end{equation}
\begin{equation}\label{eq-def-c}
  \cone = \frac{2}{(n+1.3)^2+\mucov}, \quad
  \cmu = \min\left(2\,\frac{\,\mucov-2\, +
 \frac{1}{\mucov}}{(n+2)^2 +\mucov}\,,\,1-\cone\right)
\enspace.
\end{equation}

 \subsection{Discussion of Parameters \label{sec:parameters}} The
 default values for all parameters{, namely offspring population size
 $\lambda$, recombination weights $w_{i=1,\dots,\mu}$,
 cumulation parameter $\cc$, mixing number $\mucov$, and learning
 rates $\cone$ and $\cmu$ follow
 \cite{hansen2006eda,hansen2004ppsn,hansen:08} and} were given above,
 as well as the step-size parameters test width \alphap, changing
 factor $\TPAalpha$, update bias $\TPAbeta$ and smoothing parameter
 $\calpha$.  The changes of parameters compared to
 \cite{hansen2006eda,hansen2004ppsn,hansen:08} are minor
 polishings. We discuss some settings in detail.


 \begin{description}

 \item[Recombination weights] Compared to
 \cite{hansen2006eda,hansen2004ppsn,hansen:08}, where
 $\mu'=\lceil(\lambda-1)/2\rceil$ we have chosen
 $\mu'=(\lambda-1)/2$. The small difference occurs only for even
 $\lambda$. In the former version, given odd population size
 $\lambda$, the recombination weights did not change when $\lambda$
 was reduced by one. In the present version the recombination weights
 always adjust to changes of $\lambda$.

 \item[\cone\ and \cmu] are the learning rates for the rank-one and
 rank-$\mu$ update of the covariance matrix respectively. In
 \cite{hansen2006eda,hansen2004ppsn,hansen:03,hansen:08}, a learning
 rate $\ccov \approx\cone+\cmu$ is used such that
 $\cone\approx\ccov/\mucov$ and
 $\cmu\approx\ccov(\mucov-1)/\mucov$. In the former formulation,
 $\cone$ was almost two times smaller for values of $\mucov\approx2$
 than for $\mucov=1$ and did not monotonously decrease with larger \mucov. 

 \item[\calpha] determines the smoothing of \TPAalphamean. Smoothing
 and choosing \TPAalpha\ small (damping) suppress\com{diminish}
 stochastic fluctuations of $\sigma$. In contrast to choosing 
 \TPAalpha\ small, smoothing does not affect the maximal possible change
 rate for $\sigma$. For $\calpha\ge0.5$ we find
 $\TPAalphasel\TPAalphamean>0$. Signs of the recent measurement and
 the actual change always agree and the smoothing cannot lead to
 oscillations. For $\calpha\ge0.3$ only after a second agreeing
 measure for $\TPAalphasel$ we have \emph{always}
 $\TPAalphasel\TPAalphamean>0$.  Even smaller values for \calpha\
 might be useful, but for much smaller values, presumably \TPAalpha\
 must be chosen more carefully (smaller).

%

 \item[\TPAbeta] is the bias parameter for the step-size change. On
 potentially noisy or highly rugged functions $\TPAbeta$ should be set
 to $0.2\,\TPAalpha$ which results in an effective noise handling. 
\end{description}

\section{Empirical Validation}
 In empirical investigations of the TPA-CMA-ES, we find the expected,
 feasible behavior. The comparison with CSA shows no clear
 winner. Depending on the objective function either TPA or CSA is
 faster, but the factor is seldom larger than two. Surprisingly, in
 our exploratory simulations, there is no clear winner depending on
 dimension or population size or noise. On noisy functions, setting
 $\TPAbeta=0.2\:\!\TPAalpha=0.1$\kom{+4-3 18th of quad, 3,4,5==\,\:\;} 
 for TPA is quite effective, while we
 observe only a minor effect from this change otherwise. We did not
 extensively try to exploit potential weaknesses (as has been done for
 CSA), but we suspect that the TPA is a feasible and robust
 alternative to CSA.

\section{Conclusion and Outlook}
 We see some principle \eye{advantages} of using two-point step-size
 adaptation (TPA) in the CMA-ES.

\begin{itemize} \item The TPA does not rely on a predefined
 optimality condition, like a success rate of $1/5$ or
 conjugate-perpendicularity of consecutive steps.

 \item The TPA does not rely on specific properties of the sample
 distribution or the selection of solutions. Therefore, it is supposably
 less sensitive to any modifications of the underlying algorithm, in
 particular compared to CSA.

 \item The step-size change rate can be adjusted mainly independently
 from TPA-internal considerations. Time averaging or damping are not
 essentially necessary. 
\end{itemize} 
 Even so, we see two principle \eye{disadvantages} of TPA. 

\begin{itemize} 
 \item Two additional function evaluations are needed per iteration
 step.  This is not a grave disadvantage, in particular when the
 population size is not very small. As a possible remedy, these two
 points could be incorporated in the population and used to compute
 the (final) mean in Equation \eqref{eq:meanfromx}, and one
 of them might be used in the rank-$\mu$ update of the covariance
 matrix.
\item Step-size control is based on two objective function
 evaluations only.  Selection information from the remaining
 population (and history information) is somewhat disregarded. This is
 a conceptional defect\kom{shortcoming}, that might be irrelevant in
 practice.
\end{itemize}

In conclusion, two-point step-size adaptation is an alternative to
cumulative step-size adaptation well worth of further
exploration. Whether and when it should finally replace CSA in
practice must be answered in future empirical studies. 

\bibliography{bibniko}
\bibliographystyle{plain}

\end{document}